%% file: main.tex
\documentclass[11pt,a4paper]{article}
\usepackage[hyperref]{emnlp-ijcnlp-2019}
\usepackage{times}
\usepackage{latexsym}

\usepackage{url}

\usepackage{graphicx}
\usepackage{enumitem}
\usepackage{booktabs}
\usepackage{color}
\usepackage{comment}
\usepackage{caption,subcaption}

\aclfinalcopy 

\title{Is the \emph{Red Square} Big?\\MALeViC: Modeling Adjectives Leveraging Visual Contexts}

  \author{Sandro Pezzelle \and Raquel Fern\'andez\\
Institute for Logic, Language, and Computation \\
University of Amsterdam \\
{\{\tt s.pezzelle|raquel.fernandez\}@uva.nl}}

\begin{document}

\maketitle

\begin{abstract}

This work aims at modeling how the meaning of gradable adjectives of \emph{size} (`big', `small') can be learned from visually-grounded contexts. Inspired by cognitive and linguistic evidence showing that the use of these expressions relies on setting a threshold that is dependent on a specific \emph{context}, we investigate the ability of multi-modal models in assessing whether an object is `big' or `small' in a given visual scene. In contrast with the standard computational approach that simplistically treats gradable adjectives as `fixed' attributes, we pose the problem as \emph{relational}: to be successful, a model has to consider the full visual context. By means of four main tasks, we show that state-of-the-art models (but not a relatively strong baseline) can learn the function subtending the meaning of size adjectives, though their performance is found to decrease while moving from simple to more complex tasks. Crucially, models fail in developing abstract representations of gradable adjectives that can be used compositionally.

\end{abstract}

\input{sec_introduction_new}
\input{sec_related_new}

\input{sec_tasks_new}

\input{sec_models}

\input{sec_results_new}

\input{sec_analysis_new}

\input{sec_discussion}


\section*{Acknowledgments}

We kindly thank Raffaella Bernardi and Elia Bruni for insightful discussions and their comments on an early draft of this paper, Alexander Kuhnle for sharing opinions and details regarding the models, and the audience at talks given at the ILLC and at the Blackbox@NL Workshop held at the Jheronimus Academy of Data Science. We are very grateful to the anonymous reviewers of EMNLP-IJCNLP for their valuable feedback. This work was funded by the Netherlands Organisation for Scientific Research (NWO) under VIDI grant no.~276-89-008, \emph{Asymmetry in Conversation}.


\bibliography{vague}
\bibliographystyle{acl_natbib}

\end{document}

%% file: sec_introduction_new.tex

\section{Introduction}
\label{se:intro}

There is no doubt that planets are \emph{big} things. Among the planets of our Solar System, however, Mars is unquestionably a \emph{small} planet (though not the smallest), while Saturn is definitely a \emph{big} one (though not the biggest). This example highlights some crucial properties of gradable adjectives  (hence, GAs).
First, what counts as `big' or `small' is \emph{relative}, i.e., determined by the context: Phobos is both a \emph{big} moon of Mars and a \emph{small} celestial body of the Solar System. This makes  GAs different from non-gradable or \emph{absolute} adjectives like  `open', `empty', `red': Mars, for example, is \emph{red} and \emph{rocky} in any circumstance. More formally, the compositional semantic properties of GAs are \emph{subsective} since they select a subset of entities denoted by the noun they modify, which acts as a \emph{reference set} ($\|$big$\|\subseteq\|$moon$\|$), while non-GAs are \emph{intersective} ($\|$Galilean$\|\cap\|$moon$\|$). This has consequences for the inferences they license: if Ganymede is both a \emph{Galilean} moon and a celestial body, we can infer that Ganymede is a Galilean celestial body. In contrast, if it is a \emph{big} moon and a celestial body, the inference that Ganymede is a big celestial body is not valid~\cite{partee1995lexical}.

Second, besides depending on a contextually-given reference set, GAs rely on orderings, i.e, they denote functions that map entities onto scales of degrees~\cite{cresswell1976semantics,kennedy1999projecting}. Using `big' or `small', thus, implies mapping a target object onto a size scale, which allows us to use degree morphology to express that Saturn is \emph{bigger} than Mars (comparative form) or that Mercury is the \emph{smallest} planet (superlative form). As for the non-inflected, so-called positive form of GAs (e.g., Saturn is a \emph{big} planet), its interpretation involves applying a \emph{statistical function} that makes use of a standard threshold degree~\cite{Kamp75,Pinkal79,Barker2002,Kennedy07}.

Third, GAs are considered to be \emph{vague}, because whether they apply or not to a given entity can be a matter of debate among speakers~\cite{van2012not,lassiter2017adjectival}. Since people might rely on slightly different functions involving probabilistic thresholds, there are often \emph{borderline cases}: e.g., Neptune (i.e., the fourth planet out of eight in terms of size) could be considered as a \emph{big} planet by most but not all speakers in all situations.

Our aim in this work is to computationally learn the meaning of size GAs (`big', `small') from visually-grounded contexts. 
Based on the semantic properties of such expressions (context-dependence, statistically-defined interpretation, vagueness), we tackle the task as a \emph{relational} problem in the domain of visual reasoning (similarly, e.g., to spatial problems like assessing whether `X is on the left of Y'). Simply put, a model needs to consider the entire visual context (not just the queried object) in order to solve the task. 
Such setup resembles experimental paradigms in developmental psychology which test how children interpret GAs when applied to objects grounded in visual scenes~\cite{barner2008}. Evidence shows that children learn to use GAs compositionally early on: when asked to assess whether an object is `tall' or `short' in a visual context, 4-year-old children are able to (a) restrict the reference set by means of linguistic cues and (b) derive a tall/short threshold relative to that set.  That is, 4-year-olds do not interpret GAs categorically but \emph{compositionally}. 
This is radically different from how adjectives are treated in current visual reasoning approaches, which consider them static labels standing for attributes (size, color, material) whose value is fixed across contexts~\cite{johnson2017clevr,santoro2017simple}: i.e., for current models, Saturn is always \emph{big} and Mars is always \emph{small}.

\begin{figure}[t!]
\centering
\includegraphics[width=1\linewidth]{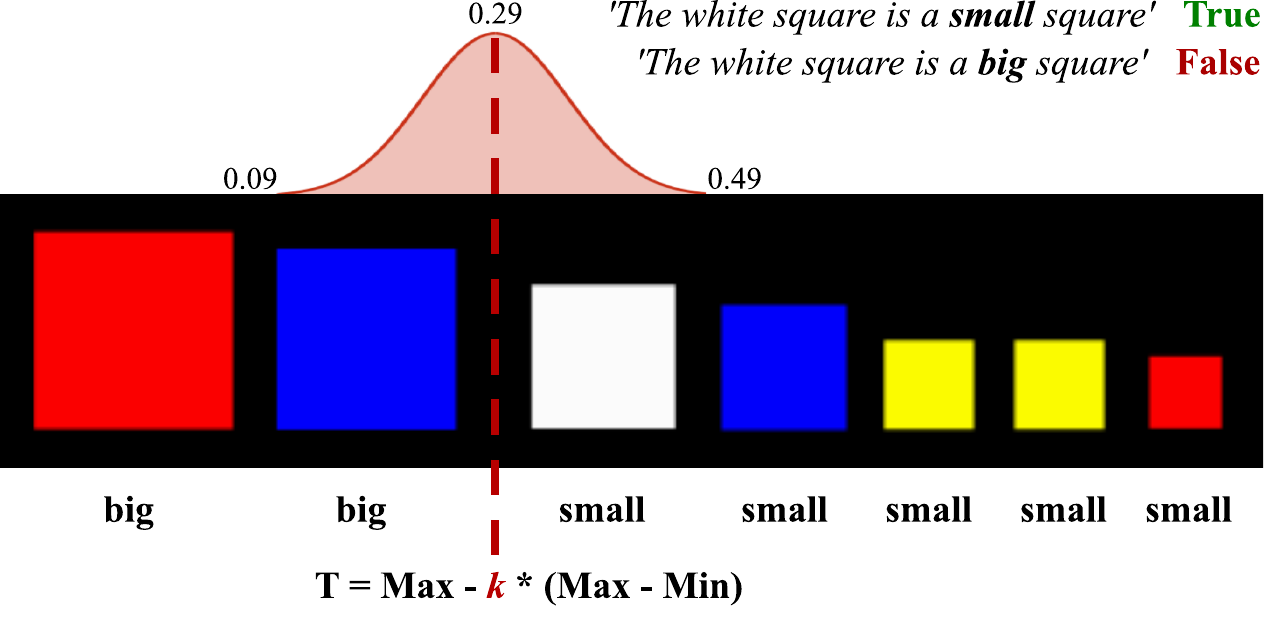}
\caption{Our approach: given a visual \textbf{context} depicting several objects and a \textbf{sentence} about one object size, models have to assess whether the sentence is true or false. Ground-truth answers are \emph{context}-specific and depend on a threshold T~\cite{SchmidtEtAl2009} which counts objects as `big' based on: (1) the area of the biggest (Max) and smallest (Min) object in the context; (2) the value of a `vague' \emph{k} determining the top \emph{k}\% of sizes which count as `big'. In our approach, all objects that are \emph{not} `big' count as `small'. Best viewed in color.}\label{fig:diagram}
\end{figure}

To model GAs in a relational fashion, we rely on a statistical function that is found to be best predictive of human interpretations~\cite{SchmidtEtAl2009} and build \textbf{MALeViC},\footnote{The name is inspired by that of the Russian 20th-century artist Kazimir Malevi\v{c} (or Malevich), famous for his paintings of geometric shapes, such as the \emph{Red Square}. Datasets, code, and trained models can be found here: \url{https://github.com/sandropezzelle/malevic}} a battery of datasets for \textbf{M}odeling \textbf{A}djectives \textbf{Le}veraging \textbf{Vi}sual \textbf{C}ontexts (see Figure~\ref{fig:diagram}). Each dataset, including 20K synthetic visual scenes and automatically-generated language descriptions (e.g., `the red square is a \emph{big} square'), is used to test different abilities. We experiment with several models and show that FiLM~\cite{perez2018film} and, to a lesser extent, Stacked Attention Networks~\cite{yang2016stacked} can learn the function subtending the meaning of size adjectives, though their performance is found to decrease while moving from simple to more complex tasks. Crucially, all models fail in developing abstract representations of gradable adjectives that can be used compositionally.

%% file: sec_related_new.tex

\section{Related Work}
\label{se:rw}

\begin{figure*}[t!]
\centering
\includegraphics[width=0.24\linewidth]{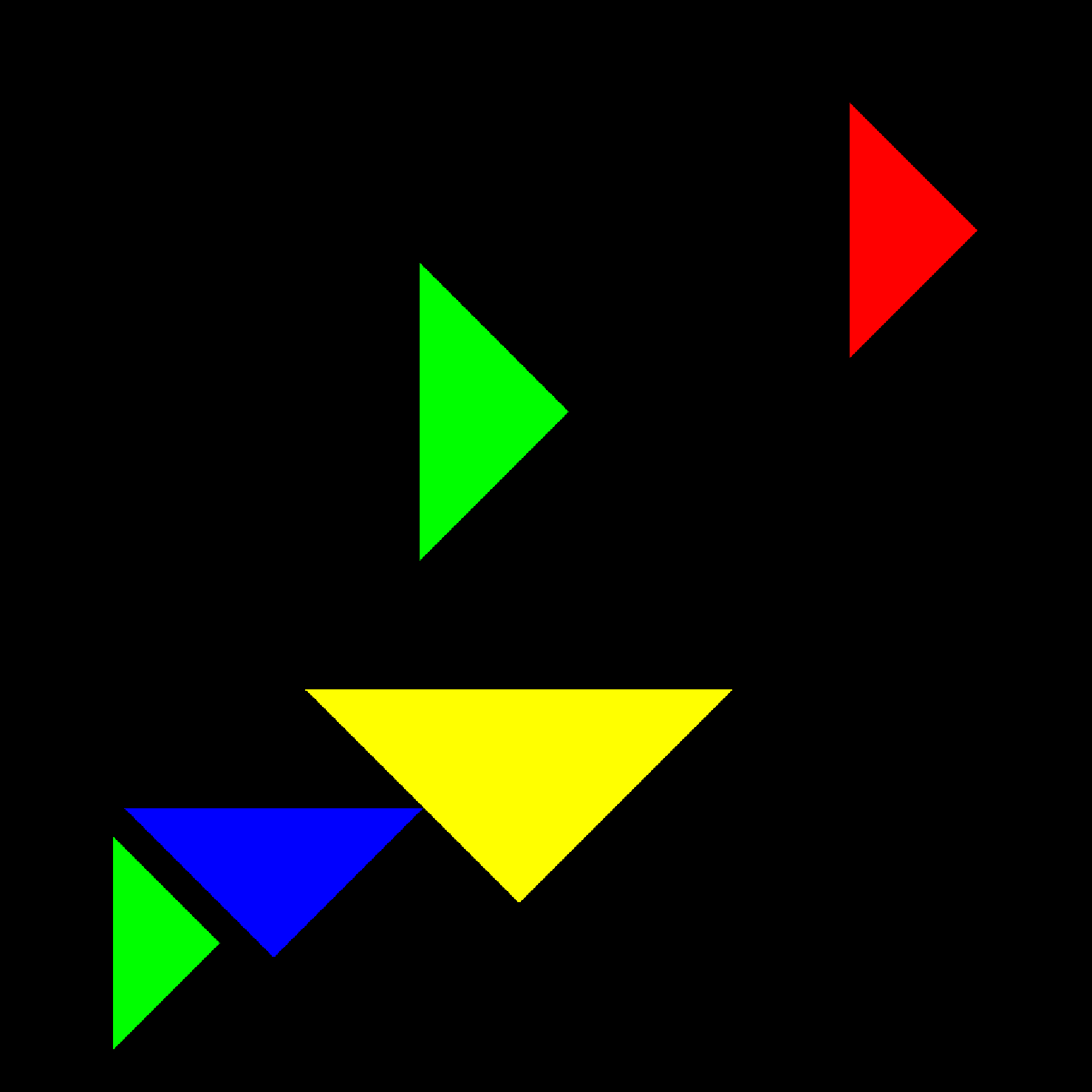}\hfill 
\includegraphics[width=0.24\linewidth]{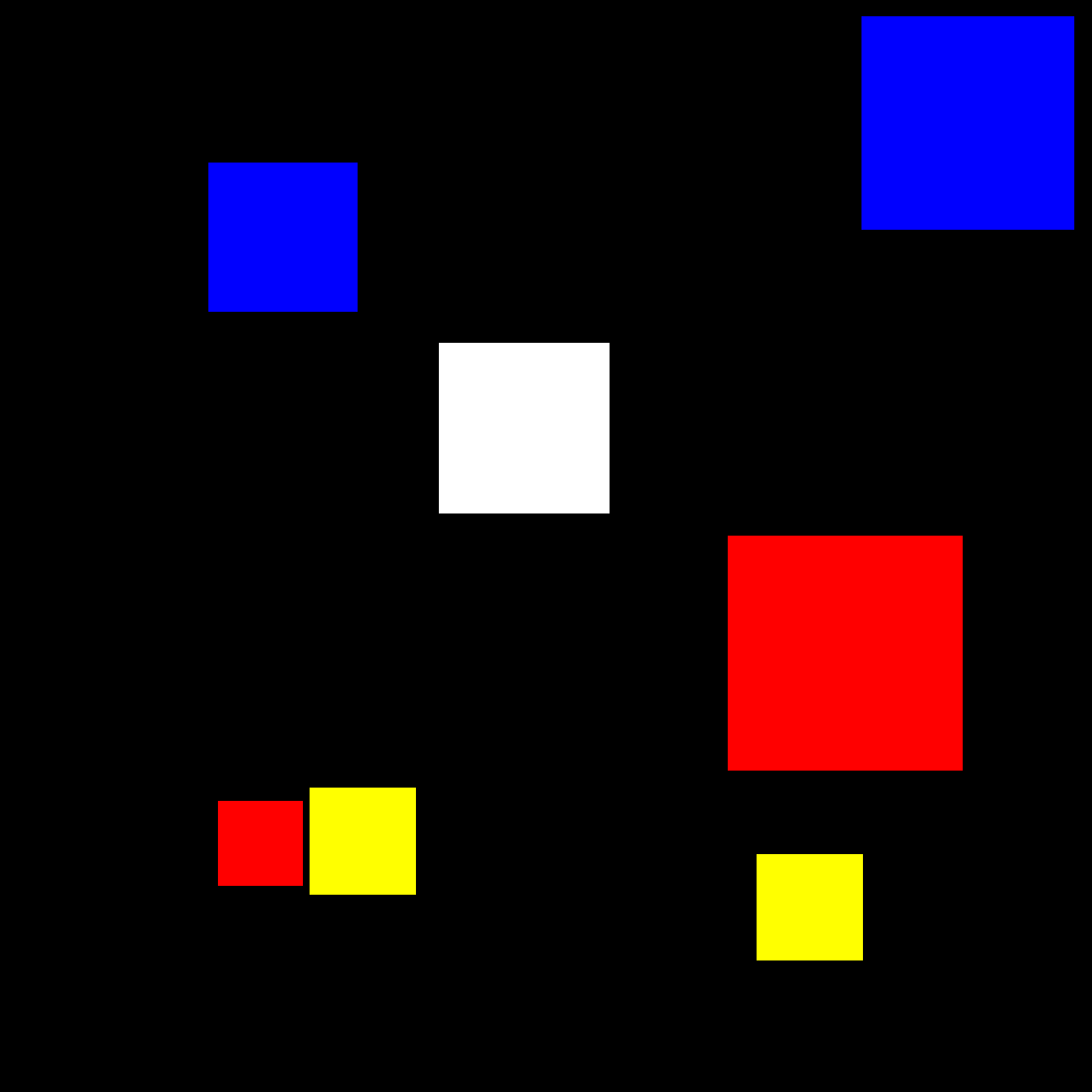}\hfill
\includegraphics[width=0.24\linewidth]{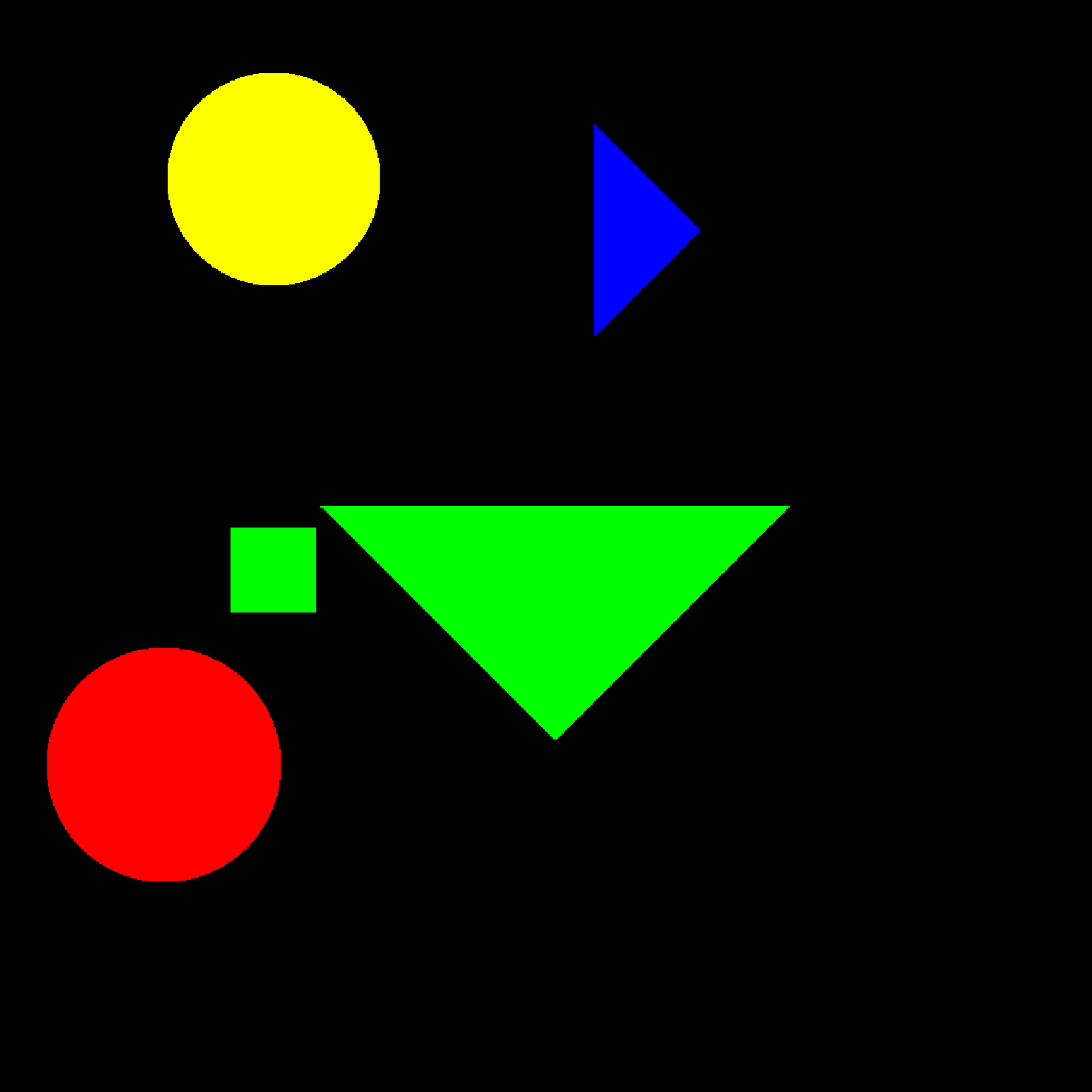}\hfill
\includegraphics[width=0.24\linewidth]{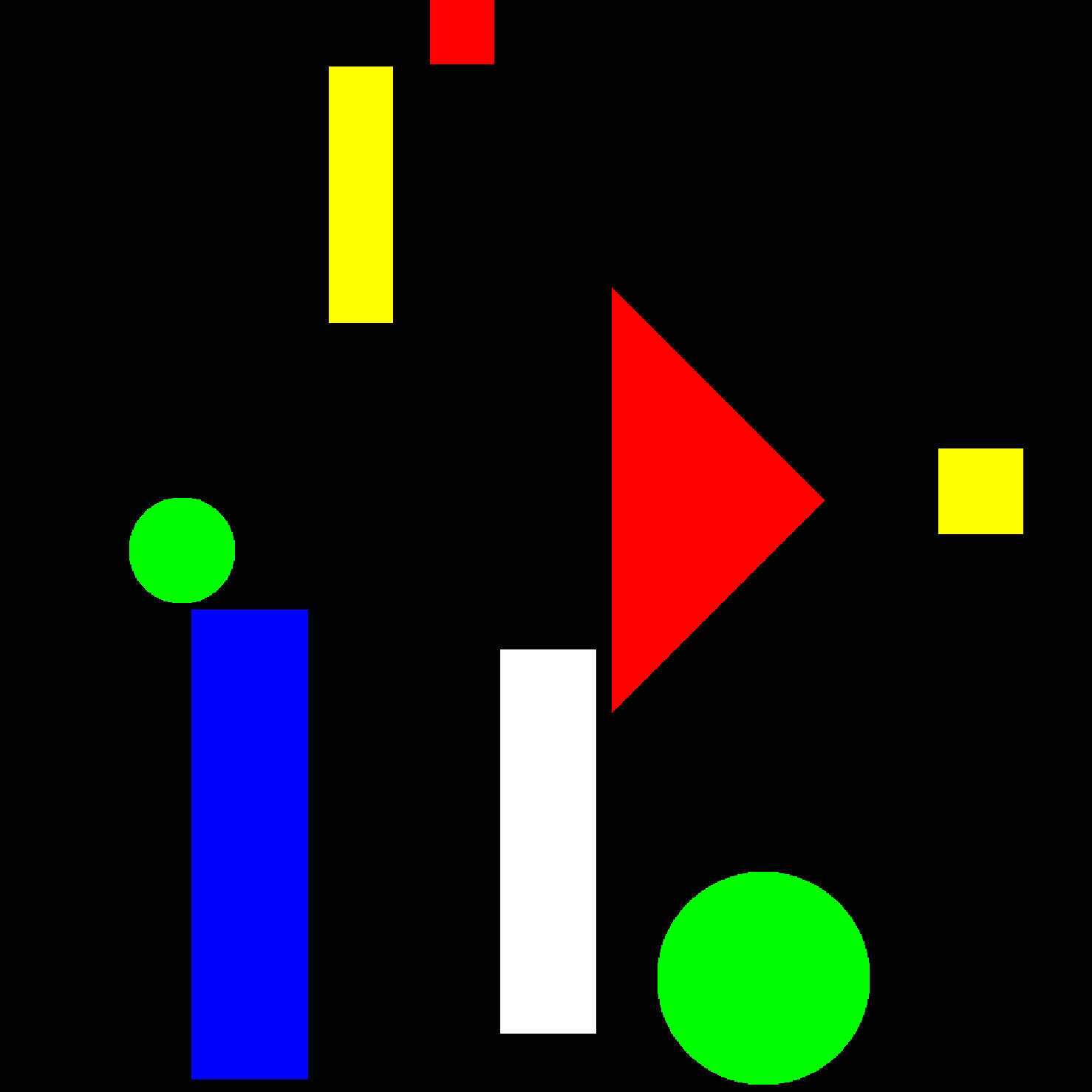}

\caption{One scene and one corresponding generated true sentence for each of the 4 tasks. From left to right, \textbf{SUP\textsubscript{1}:} The yellow triangle is the \emph{biggest} triangle; \textbf{POS\textsubscript{1}:}  The white square is a \emph{small} square; \textbf{POS:}  The red circle is a \emph{big} object; \textbf{SET+POS:} The white rectangle is a \emph{big} rectangle. Best viewed in color.}\label{fig:examples}
\end{figure*}

\paragraph{Computational Linguistics} 
Computational approaches to GAs have mostly focused on automatically ordering elements with respect to their intensity~\cite[e.g., \emph{good}$<$\emph{great}$<$\emph{excellent},][]{de2010good}
to overcome a problem with lexical resources like WordNet~\cite{Fellbaum1998}, 
which consider words like `small' and `minuscule' as synonyms.
These efforts showed the potential of using techniques based on word embeddings~\cite{kim2013deriving}, web-scale data~\cite{sheinman2013large,de2013good}, or their combination~\cite{shivade2015corpus,kim2016adjusting} to determine the relative intensity of different words on a scale. 
By focusing on the \emph{ordering} between adjectives, however, these works do not shed light on how the \emph{meaning} of such expressions is determined by contextual information. This goal, in contrast, has been pursued by work on automatic Generation of Referring Expressions (GRE), where GAs have represented an interesting test case precisely due to their context-dependent status~\cite{vanDeemter06}. Several small datasets of simple visual scenes and corresponding descriptions have been proposed~\cite{van2006building,viethen2008use,viethen2011gre3d7,mitchell2010natural,mitchell2013typicality}, and forerunner GRE algorithms have been tested to refer to visible objects using attributes like \emph{size}~\cite{mitchell2011applying,mitchell2013generating}. Due to their extremely small size, however, these datasets are not suitable for large-scale deep learning investigations.

\paragraph{Language \& Vision} 
Recent work in language and vision has dealt with (gradable) adjectives from at least three perspectives. First, multi-modal information has been used to order real-world entities with respect to an attribute described by a GA.~\citet{bagherinezhad2016elephants}, for example, assess whether elephants are \emph{bigger} than butterflies. Thus, rather than modeling the meaning of GAs, this work focuses on the relative size of object types, which is crucially different from our goal.
Second, visual gradable attributes have been used as fine-grained features to answer questions about entities in natural scenes~\cite{antol2015vqa,hudson2019gqa}, to perform discriminative and pragmatically-informative image captioning~\cite{Vedantam_2017_CVPR,cohn2018pragmatically}, and to discriminate between similar images in reference games~\cite{su2017reasoning}. In these works, however, GAs are labels standing for \emph{fixed} attributes: e.g., an airplane is annotated as `large' or `small' with no or little relation to the visual context. Third, gradable attributes are explored in work on visual reasoning~\cite{johnson2017clevr,yi2018neural}, where multi-modal models are challenged to answer complex, high-level questions about objects and their relations in synthetic scenes. Surprisingly, however, questions involving GAs are treated as \emph{non-relational}: e.g., a given size \emph{always} corresponds to either `big' or `small', regardless of the objects in the visual context. One exception is~\citet{kuhnle2018clever}, whose investigation of superlative (`biggest') and comparative (`darker') GAs represents a first step toward a \emph{relational} account of these expressions. To our knowledge, we are the first to tackle the modeling of \emph{positive} GAs as a linguistically- and cognitively-inspired relational problem in language and vision.

%% file: sec_tasks_new.tex

\section{Tasks}
\label{se:tasks}

To test model abilities in learning size GAs, we propose 4 main tasks (see Figure~\ref{fig:examples}). All tasks are framed as a sentence-verification problem: given an image depicting colored, geometric shapes and a sentence about one object's relative size, systems are asked to verify whether the given sentence is true or false in that context. These tasks are aimed at assessing one ability at a time, and are specifically designed to form a pipeline of increasingly difficult operations. 
Tasks differ with respect to:

\begin{itemize}[itemsep=-1pt,leftmargin=11pt]
\item the \textbf{statistical function} at play: a max/min function (SUP) or a threshold function (POS);
\item the number of geometric \textbf{shape types}: one  (SUP\textsubscript{1}/POS\textsubscript{1}) or several;
\item the scope of the \textbf{reference set}: the entire visual scene or a subset of objects (SET; only applicable with several shape types per scene).
\end{itemize}

\paragraph{SUP\textsubscript{1}}
This task tests whether a model is able to interpret size GAs in their \textbf{superlative} form: e.g., `The yellow triangle is the \emph{biggest} triangle' (only triangles in the scene).
To solve this task, a model is required to (1) identify the queried object, (2) measure object size, and (3) determine whether the target object has the largest or smallest size in the entire visual scene.

\paragraph{POS\textsubscript{1}}
This task evaluates the ability of models to interpret \textbf{positive}-form size GAs: e.g., `The white square is a \emph{small} square' (only squares in the scene).
To adequately solve this task, a model is not only required to (1) identify the queried object and (2) measure object size, but also (3)  to learn the threshold function that makes an object count as `big' or `small' depending on the size of the other objects in the entire visual scene, and (4) to assess whether the target object counts as `big'/`small' in this context. In contrast to the superlative form, which is precise, the positive form is \emph{vague}: there may be borderline cases (see Figure~\ref{fig:diagram} and \emph{Threshold Function} in Section~\ref{se:datasets}).

\paragraph{POS}
This task is an extension of POS\textsubscript{1} where the restriction to one single geometric shape type per scene is lifted:  e.g., `The red circle is a \emph{big} object' (any shape in the scene) -- see Figure~\ref{fig:examples}. As before, and in contrast to the next task, the reference set is the entire visual scene.

\paragraph{SET+POS}
Finally, this task assesses the ability of models to interpret \textbf{positive}-form GAs with respect to a restricted context: e.g., `The white rectangle is a \emph{big} rectangle' (any shape in the scene). To solve this task, in addition to the skills required to address the POS task, a model needs to determine the relevant \textbf{reference set} (e.g., the set of rectangles in the scene) and to apply all  POS operations to this set. This task, thus, brings together all the key components that make up the semantics of size adjectives, as described in Section~\ref{se:intro}.


\section{Datasets}
\label{se:datasets}

\paragraph{Visual Data} 
For each task, we build a dataset of synthetic scenes (1478 x 1478 pixels) depicting 5 to 9 colored objects\footnote{This resembles the setup used by \citet{barner2008} in their first experiment, which involves 9 objects.} on a black background. Each object is randomly assigned one shape among circle, rectangle, square, triangle; one color among red, blue, white, yellow, green; one area among 10 predefined ones (based on number of pixels) that we label using tens ranging from 30 to 120 (i.e. 30, 40, 50, \dots 120); and a given spatial position in the scene which avoids overlapping with other objects. In tasks SUP\textsubscript{1} and  POS\textsubscript{1} only one shape type is present in a given scene, while in tasks POS and SET+POS several shape types are present.

\begin{figure}[t!]
\centering
\includegraphics[width=1\linewidth]{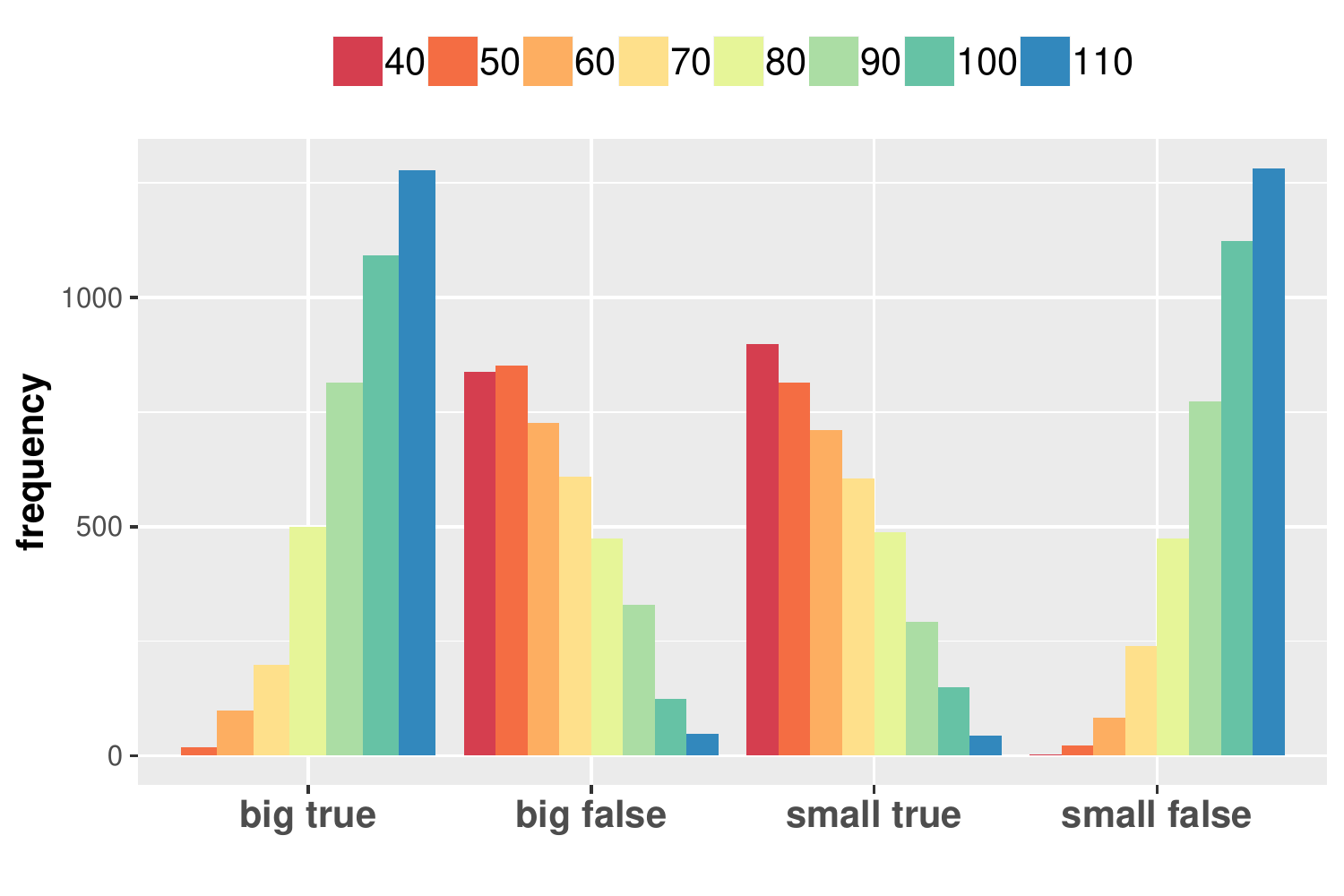}
\caption{\textbf{POS.} Distribution of queried objects' areas per question type in train. This plot highlights an important feature of MALeViC, namely that objects with any area from 40 to 110 can be either `big' or `small' depending on the visual context. Best viewed in color.}
\label{fig:areas}
\end{figure}

\paragraph{Linguistic Data}
While generating the visual data, ground-truth labels (`biggest', `smallest', `big', `small') are automatically assigned to each object based on the area of the objects present in the reference set. For tasks POS\textsubscript{1},  POS and SET+POS, for each object in the scene the generation algorithm generates a sentence based on the template: \emph{The} $<$color$>$ $<$shape$>$ \emph{is a} $<$size$>$ $<$shape$>$. For task SUP\textsubscript{1}, for objects that are biggest/smallest in the scene the algorithm generates a sentence using the template \emph{The} $<$color$>$ $<$shape$>$ \emph{is the} $<$sup\_size$>$ $<$shape$>$. In order for a sentence to be licensed, the following constraints have to be met: (1) The queried object is uniquely identifiable within the scene, either by $<$color$>$ in tasks SUP\textsubscript{1} and POS\textsubscript{1} or by $<$color, shape$>$ in tasks POS and SET+POS (this ensures there is no ambiguity regarding the target); (2) the area of the queried object must be in the 40-110 range, so to avoid querying objects whose size would always be small/smallest or big/biggest in any scene irrespective of the area of the other objects (see Figure~\ref{fig:areas}, which shows the distribution of queried objects' areas per question type in the train set of POS; as can be seen, no objects with an area equal to either 30 or 120 are present). In task SET+POS, we include two additional constraints: (3) the queried object is neither the biggest nor the smallest object in the entire scene (i.e., the max/min function is not effective without identifying the reference subset); and (4) there are at least three objects in the scene with the same shape as the target (i.e., the reference set includes at least three objects), so to make the computation of the threshold function required to solve the task. Only sentences meeting these constraints are generated and the corresponding scene retained. For each true sentence, a false sentence is automatically generated by replacing the true adjective (e.g., `big') with the false one (e.g., `small').

\paragraph{Threshold Function}
To assign ground-truth labels to objects in tasks POS\textsubscript{1}, POS and SET+POS, we make use of a \emph{threshold function} experimentally determined by~\citet{SchmidtEtAl2009}, who explore a number of statistical functions to describe speakers' use of the adjective `tall' (following the authors, we assume these functions to be valid for any GAs, including `big').\footnote{Note also that, while size may be argued to be two-dimensional, in our approach we treat it as one-dimensional (i.e., based on number of pixels), similarly to tallness.} In particular, we use their \emph{Relative Height by Range} (RH-R) function, according to which any item within the top $k\%$ of the range of sizes in the reference set is `tall' (`big'). According to this function, the threshold T for a given context C is defined as follows: T(C) = Max - \emph{k} * (Max -- Min), where $\emph{k} \in [0,1]$, Max is the maximum size in C, and Min the minimum size (see Figure~\ref{fig:diagram}). 
In our data, we make the simplifying assumption to consider `small' any object that is \emph{not} `big'. This way, we also avoid dealing with negative statements. Since~\citet{SchmidtEtAl2009} experimentally show \emph{k} = 0.29 to best fit their human data, we use this value as our reference \emph{k}. 
To account for \emph{vagueness}, for each scene we randomly sample a \emph{k} from the normal distribution of values centered on 0.29 ($\mu$ = 0.29, $\sigma$ = 0.066),\footnote{By setting $\sigma$ = 0.066, we expect 99.7\% \emph{k} values to be $\pm$ 3 standard deviations from 0.29, i.e. ranging from 0.09 to 0.49; 95\% \emph{k} within $\pm$ 2 SD (0.16-0.42); 68\% within $\pm$ 1 SD (0.22-0.36). $\sigma$ value was set experimentally to keep \emph{k} $<$ 0.5.} as illustrated in Figure~\ref{fig:diagram}. This introduces some perturbation in the definition of T, which mimics the fact that speakers may rely on slightly different definitions of GAs \cite[i.e., what counts as `big' for one person might not \emph{always} count as `big' for another one;][]{raffman1996vagueness,shapiro2006vagueness,sharp2018grounding}, with communication still being successful in most of the cases.

\paragraph{Datasets} 
The four final datasets, including 20K datapoints each, are perfectly balanced with respect to the number of cases for each combination of variables used. In particular, 250 $<$scene, sentence$>$ datapoints are included for each of the 80 classes (4 shapes * 5 colors * 2 sizes * 2 ground truths), where one class is e.g.: $<$red$>$ $<$circle$>$ $<$big$>$ $<$true$>$. Such balancing is kept when randomly splitting the datasets into train (16K cases), validation (2K) and test (2K) sets.

%% file: sec_models.tex
\section{Models}
\label{se:models}

We test 3 models that have proved effective in visual reasoning tasks~\cite{johnson2017clevr,suhr2018corpus,yi2018neural}. All models are \emph{multi-modal}, i.e., they use both a visual representation of the scene and a linguistic representation of the sentence. Unless otherwise specified, visual features representing the scenes are extracted via a \emph{fixed} Convolutional Neural Network (CNN) fed with images resized to 224 x 224 pixels. In particular, features from the \emph{conv4} layer of a ResNet-101~\cite{he2016deep} pre-trained on ImageNet~\cite{russakovsky2015imagenet} are used.

\paragraph{CNN+LSTM} This model simply  concatenates the CNN visual features with a representation of the sentence encoded using the final hidden state of an LSTM~\cite{hochreiter1997long}. These concatenated features are passed to a Multi-Layer Perceptron (MLP) that predicts the answer.

\paragraph{Stacked Attention (CNN+LSTM+SA)} This model combines the CNN visual features and the LSTM final state by means of two rounds of soft spatial attention. An MLP over the combined representation predicts the answer~\cite{yang2016stacked}.

\paragraph{CNN+GRU+Feature-wise Linear Modulation (FiLM)} This model~\cite[$\sim$100\% on CLEVR;][]{johnson2017clevr} processes the image features by means of 4 residual blocks where the visual representation is linearly transformed by the sentence embedding~\cite[i.e., the final hidden state of a GRU;][]{chung2014empirical}. After a global max-pooling, a two-layer MLP classifier outputs a softmax distribution over the answers~\cite{perez2018film}.\footnote{For CNN+LSTM and CNN+LSTM+SA, we use the implementations by~\citet{johnson2017inferring}. For FiLM, we use the implementation by~\citet{perez2018film}. All models are trained using Python 3.5.2 and PyTorch v1.0.1.}

\paragraph{Experimental Setup} For each model in each task, we experiment with 16 configurations of hyper-parameters, i.e., 4 learning rates (5e-5, 3e-4, 5e-4, 5e-3) * 2 dropout values (0, 0.5) * 2 batch normalization options (yes, no). Each model configuration is trained 3 times with randomly initialized weights for 10K iterations (40 epochs), and the best configuration is chosen based on average validation accuracy. In total, 576 models (3 architectures * 4 tasks * 48 runs) are tested. For comparison with previous work~\cite{perez2018film,kuhnle2018clever}, we also test FiLM end-to-end (i.e., trained from raw pixels) with the same hyper-parameters as above, for a total of 192 models (4 tasks * 48 runs). Since no substantial differences are observed between the two versions~\cite[in line with][]{perez2018film}, we focus on the results by the less tailored and less computationally-expensive models using \emph{pre-trained} features.

%% file: sec_results_new.tex

\section{Results}
\label{se:results}

\begin{table*}[t!]
\small
\resizebox{\textwidth}{!}{
\begin{tabular}{|l|l|l|l|l|lll|}
\hline
\multicolumn{1}{|c|}{\textbf{task}} & \multicolumn{1}{c|}{\textbf{model}} & \multicolumn{3}{c|}{\textbf{accuracy}}                                                                                & \multicolumn{3}{c|}{\textbf{hyper-parameters}} \\ \hline
\multicolumn{1}{|c|}{\textbf{}}     & \multicolumn{1}{c|}{\textbf{}}      & \multicolumn{1}{c|}{max val (pixels)} & \multicolumn{1}{c|}{avg val $\pm$ sd} & \multicolumn{1}{c|}{avg test $\pm$ sd} & lr             & drop          & b norm           \\ \hline
                                    & CNN-LSTM                            & 0.841                                 & 0.8153 $\pm$ 0.033                   & 0.8066 $\pm$ 0.033                     & 5e-4           & 0.5           & no           \\
\textbf{SUP\textsubscript{1}}                          & CNN-LSTM-SA                         & \textbf{1}                            & 0.999 $\pm$ 0                        & 0.9983 $\pm$ 0.001                     & 3e-4           & 0            & no           \\
                                    & FilM                                & \textbf{1 (1)}                        & \textbf{0.9991 $\pm$ 0}              & \textbf{0.999 $\pm$ 0.001}             & 3e-4           & 0.5           & no           \\ \hline \hline   

                                    & CNN-LSTM                            & 0.5615                                & 0.5493 $\pm$ 0.009                   & 0.5455 $\pm$ 0.013                     & 5e-4           & 0.5           & yes          \\
\textbf{POS\textsubscript{1}}                          & CNN-LSTM-SA                         & 0.941                                 & \textbf{0.9396 $\pm$ 0.001}          & \textbf{0.9306 $\pm$ 0.002}            & 3e-4           & 0            & no           \\
                                    & FiLM                                & \textbf{0.9415 (0.9565)}              & 0.8673 $\pm$ 0.085                   & 0.8546 $\pm$ 0.086                     & 5e-5           & 0.5           & no           \\ \hline \hline 

                                    & CNN-LSTM                            & 0.574                                 & 0.5668 $\pm$ 0.005                   & 0.5493 $\pm$ 0.008                     & 3e-4           & 0.5           & yes          \\
\textbf{POS}                          & CNN-LSTM-SA                         & 0.942                                 & \textbf{0.9386 $\pm$ 0.002}          & \textbf{0.94 $\pm$ 0.002}              & 3e-4           & 0            & yes          \\
                                    & FiLM                                & \textbf{0.945 (0.9475)}               & 0.9375 $\pm$ 0.004                   & 0.9333 $\pm$ 0.002                     & 5e-4           & 0            & no           \\ \hline \hline 
                                    & CNN-LSTM                            & 0.591                                 & 0.5808 $\pm$ 0.014                   & 0.551 $\pm$ 0.01                       & 5e-4           & 0.5           & yes          \\
\textbf{SET+POS}                          & CNN-LSTM-SA                         & 0.81                                  & 0.7901 $\pm$ 0.014                   & 0.7751 $\pm$ 0.01                      & 5e-5           & 0            & no           \\
                                    & FiLM                                & \textbf{0.9205 (0.9295)}              & \textbf{0.8845 $\pm$ 0.027}          & \textbf{0.8788 $\pm$ 0.021}            & 5e-4           & 0            & no           \\ \hline\hline 
	\textbf{\emph{all}}				& \textit{chance}                & 0.5                                & 0.5                               & 0.5                                 &                &               & \\ \hline

\end{tabular}
}
\caption{Results by each \textbf{model} (column 2) in each \textbf{task} (1). Note that \textbf{max val} (3) refers to the highest accuracy obtained in the task by a given architecture across 192 runs (in brackets, highest accuracy obtained across 192 runs by FiLM trained from raw \textbf{pixels}), while \textbf{avg val $\pm$ sd} (4) and \textbf{avg test $\pm$ sd} (5) refer to the average accuracy (and relative standard deviation) obtained across 3 runs by the best configuration of hyper-parameters of a given architecture on, respectively, val and test set. As for the \textbf{hyper-parameters} (6), we report value of learning rate (\textbf{lr}), dropout (\textbf{drop}), and use/not use of batch normalization (\textbf{b norm}). Values in bold are the highest in the column.}\label{tab:results}
\end{table*}

Overall, we observe the expected pattern of results, in line with the conjectured, increasing difficulty of the tasks: accuracy declines from SUP\textsubscript{1} to POS\textsubscript{1} and from POS to SET+POS. There are, however, clear differences across models. CNN+LSTM does well on SUP (the simplest task) but performs around chance on all other tasks (which require a threshold function). Both CNN+LSTM+SA and FiLM obtain high accuracy on POS\textsubscript{1}/POS, while FiLM neatly outperforms CNN+LSTM+SA in the most challenging SET+POS task, proving its higher ability to reason over complex, context-dependent linguistic expressions.
Accuracy scores are reported in Table~\ref{tab:results}. In the following, we describe the results per task in more detail.

\paragraph{SUP\textsubscript{1}} 
Both CNN+LSTM+SA and FiLM obtain perfect accuracy ($\sim$100\%), with CNN+LSTM performing well above chance ($\sim$80\%). These results indicate that assessing whether an object is the `biggest'/`smallest' in the scene is a relatively simple task even for basic models. This is encouraging since mastering \emph{superlatives} requires several core skills that also underlie the use of \emph{positive} GAs, namely (1) object identification, (2) object size estimation, and (3) object size ordering. 

\paragraph{POS\textsubscript{1} / POS} 
CNN+LSTM+SA and FiLM obtain similar, very high max accuracy ($\sim$94\%) in these two tasks, with CNN+LSTM being only slightly above chance level ($\sim$57\%). Note that in this case we should not expect models to obtain 100\% accuracy: while the min/max function in SUP is precise, the POS threshold function is vague (giving rise to borderline cases). The performance to be expected with sharp $k$ = 0.29 is 97\% for these two tasks (i.e, only in 3\% of cases $k$ is assigned a value that makes the truth/falsity of a sentence different from what it would be with $k$ = 0.29). The performance of the two top models is thus 3\% below average ceiling accuracy (94\% vs.~97\%).
These results, thus, indicate that CNN+LSTM+SA and FiLM are able to learn the \emph{threshold function} that makes an object count as `big'/`small' in a given visual scene, while CNN+LSTM is not.\footnote{FiLM's performance in POS\textsubscript{1} (where each scene contains only one shape type, but all shape types are seen across images) is rather unstable across runs (high SD). Since this is not so in POS, we conjecture that it may be due to the model learning shape-specific strategies in some runs of POS\textsubscript{1}.}

\paragraph{SET+POS}
As before, CNN+LSTM performs only slightly above chance. Both FiLM and CNN+LSTM+SA experience a drop in performance compared to 
POS\textsubscript{1}/POS, confirming that computing the threshold function over a subset of objects is the most challenging task. Applying the threshold function to the entire visual scene would yield $\sim$65\% accuracy in this dataset. This indicates that none of the two top models uses this strategy simplistically, as their performance is well above this result (max 81\% for CNN+LSTM+SA and 92\% for FiLM).\footnote{As in POS\textsubscript{1}/POS, ceiling performance with fixed $k$ = 0.29 is 97\% in this dataset.}   
FiLM neatly outperforms CNN+LSTM+SA (+11\% in both max and average accuracy), thus showing a more robust pattern of results across tasks and a higher ability to handle complex reasoning problems.

While these results are very encouraging, they might be the outcome of exploiting biases in the data rather than learning the semantic components that make up the meaning of GAs according to linguistic theory and psycholinguistic evidence. For example, identifying the reference subset but applying the max/min function (rather than a threshold function) to this set would yield a remarkable 92\% accuracy, which is on a par with the top result by FiLM. 
In what follows, we investigate the abilities of the trained models in more depth by testing them on a number of diagnostic test sets.

%% file: sec_analysis_new.tex

\section{Analysis}
\label{se:analysis}

In this section, we aim at better understanding what the models learn (or do not learn) when performing the different tasks. To do so, we carry out a \emph{stress-test} and a \emph{compositionality} analysis. 

\subsection{Stress-Test: Hard Contexts}
\label{subsec:hard}

To test to what extent models master the core abilities that are arguably needed to perform POS and SET+POS, we build two \emph{hard} diagnostic test sets which make the use of other strategies not effective. Similarly to the main datasets, these hard test sets include 2K perfectly-balanced datapoints. 

\paragraph{POS-hard} 

\begin{table}[t!]
\small
\resizebox{\columnwidth}{!}{
\begin{tabular}{|l|l|l|}
\hline
\multicolumn{1}{|c|}{\textbf{hard test (train)}} & \multicolumn{1}{c|}{\textbf{model}} & \multicolumn{1}{c|}{\textbf{avg acc $\pm$ sd}} \\ \hline
\textbf{POS-hard}                                     & CNN+LSTM                                    & 0.5325 $\pm$ 0.01                               \\
\textbf{(POS)}                           & CNN+LSTM+SA                               & 0.8653 $\pm$ 0.005                              \\
                                     & FiLM                                      & \textbf{0.8693 $\pm$ 0.003}                     \\ \hline\hline 
\textbf{SET+POS-hard}                                     & CNN+LSTM                               & 0.4623 $\pm$ 0.004                              \\
\textbf{(SET+POS)}                       & CNN+LSTM+SA                            & 0.478 $\pm$ 0.015                               \\
                                     & FiLM                                   & \textbf{0.6513 $\pm$ 0.059}                     \\ \hline \hline 
		\textbf{\emph{all}}			& \emph{chance}	& 0.5 \\ \hline

\end{tabular}
}
\caption{\textbf{Average accuracy} across 3 runs by a \textbf{model} in a \textbf{hard test} set. Models are trained on the task in \textbf{train}. Values in bold are the highest in the column.}
\label{tab:hard_tasks} 
\end{table}

We explore whether models trained on POS properly learn to compute the threshold function rather than using a simplistic strategy such as applying the min/max function over the visual context. 
Indeed, it has been proposed that young children might use positive-form GAs as superlatives in early stages of language acquisition~\cite{clark1970primitive}.
To test whether this explains (part of) the results on POS, we build a \emph{hard} test set identical to POS except for the fact that objects that are either the biggest or the smallest in the entire visual scene are never queried. Thus, in POS-hard systematically using the superlative strategy would result in chance level scores. As reported in Table~\ref{tab:hard_tasks}, the accuracy obtained by CNN+LSTM+SA and FiLM does not dramatically decrease compared to POS ($\sim$94\% vs.~$\sim$87\%). The drop can be explained by the fact that in POS-hard we query objects that are overall \emph{closer} to the threshold, thus increasing the number of more difficult, borderline cases.\footnote{This also leads to slightly lower average ceiling accuracy in POS-hard: performance with fixed $k$ = 0.29 is now $\sim$92\% in contrast to $\sim$97\% in POS.} 
This test thus confirms that CNN+LSTM+SA and FiLM do learn to generally compute and apply the threshold function over an entire visual scene.

\begin{figure}[t!]
\centering
\includegraphics[width=1\linewidth]{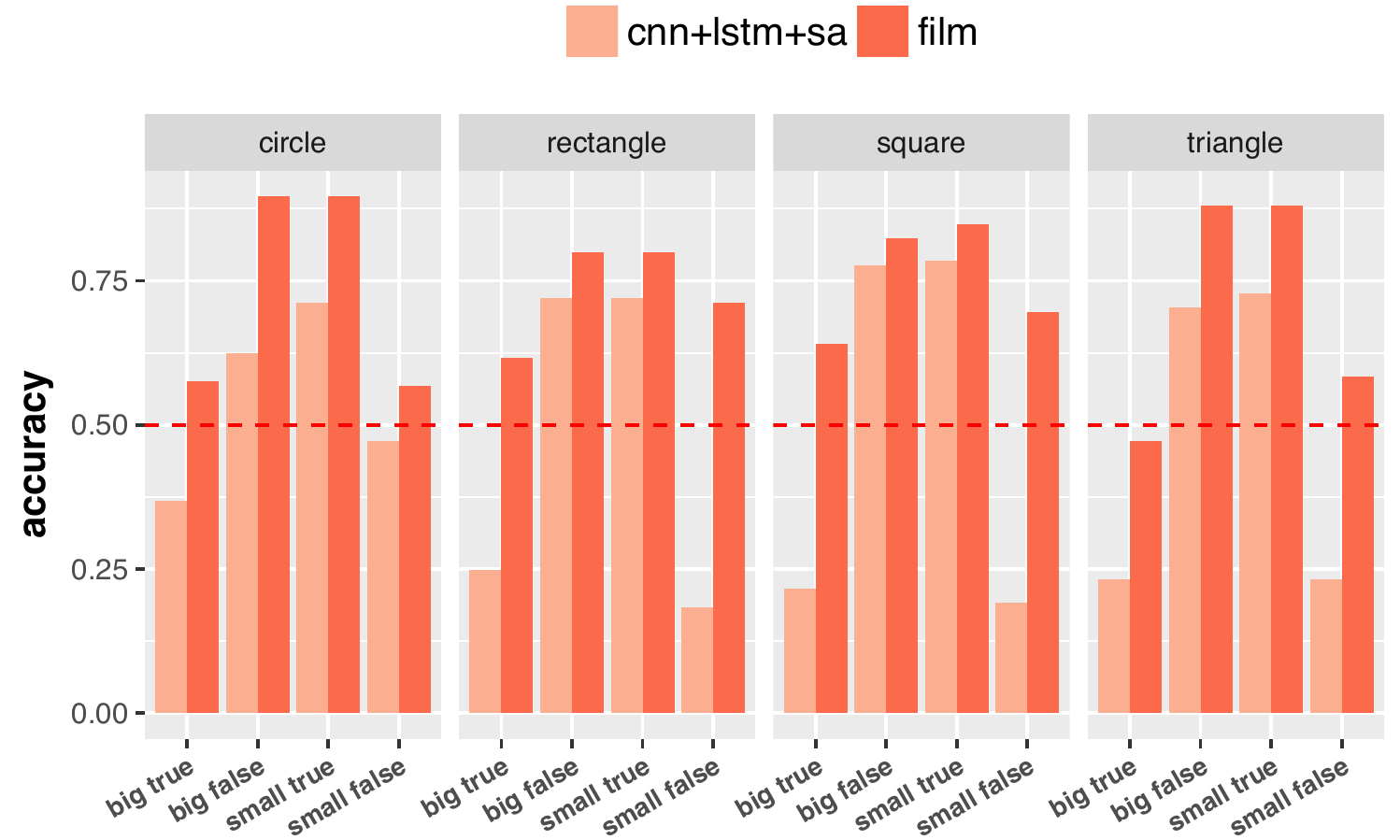}
\caption{\textbf{SET+POS-hard.} Accuracy per sentence type (big true, big false, small true, small false) obtained by the best run of CNN+LSTM+SA (49\% acc.) and FiLM (73\% acc.) for each shape type. The dashed line signals \emph{chance} level. Best viewed in color.}\label{fig:dhard_acc_shape} 
\end{figure}

\paragraph{SET+POS-hard} 
To investigate whether models trained on SET+POS learn to identify the reference subset and to apply the threshold function to it, we build a hard test set identical to 
 SET+POS except for the fact that objects that are the biggest or the smallest \emph{in the reference set} are never queried. Thus, 
 in SET+POS-hard applying the min/max function to the reference set 
 would lead to chance level. 
 As shown in Table~\ref{tab:hard_tasks}, only FiLM is above chance level in this test set, with 65\% accuracy. Both CNN+LSTM and CNN+LSTM+SA obtain scores slightly below chance (--30\% for  CNN+LSTM+SA compared to SET+POS). This is a striking result, which reveals that the accuracy scores achieved by CNN+LSTM+SA in SET+POS must be due to shortcut strategies.

While we do not have full understanding of what is being exploited by the CNN+LSTM+SA model, we do observe a bias towards predicting that an object counts as `small'. As shown in Figure~\ref{fig:dhard_acc_shape}, the model obtains high accuracy on sentences involving small objects (small true, big false), while its accuracy on sentences targeting big objects (big true, small false) is below chance level. Indeed, the model predicts \emph{false} for big objects and \emph{true} for small objects in 73\% of cases.

\begin{figure}[t!]
\centering
\includegraphics[width=1\linewidth]{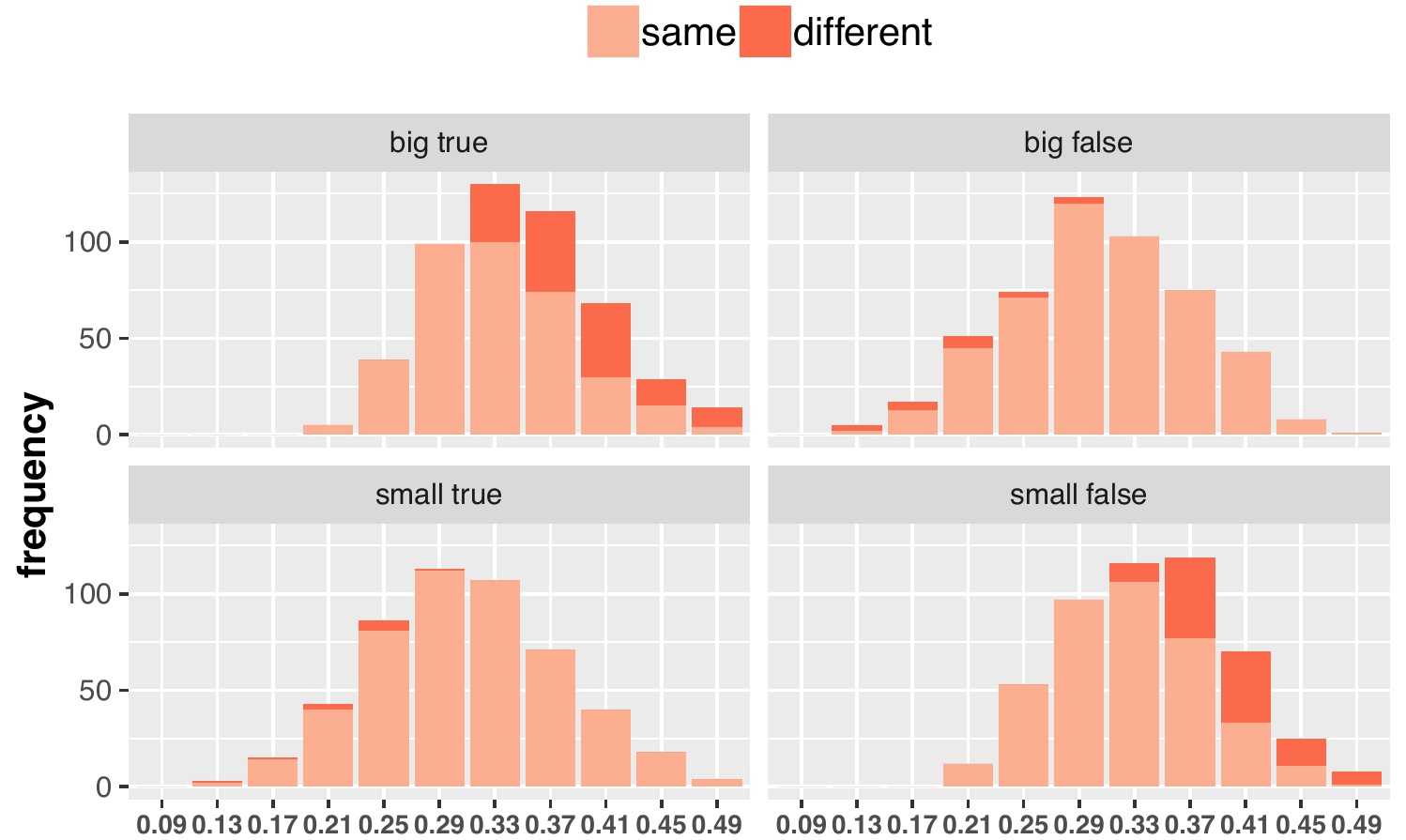}
\caption{\textbf{SET+POS-hard.} Distribution of \emph{k} against sentence type (clockwise from top-left: big true, big false, small false, small true). Cases labelled as \emph{different} are those for which the ground truth adjective (e.g. `big') would change with \emph{k} = 0.29 (e.g., `small'); in \emph{same} cases, it would be the same. Best viewed in color.}\label{fig:distr_k_dhard} 
\end{figure}

\begin{figure}[t!]
\centering
\includegraphics[width=1\linewidth]{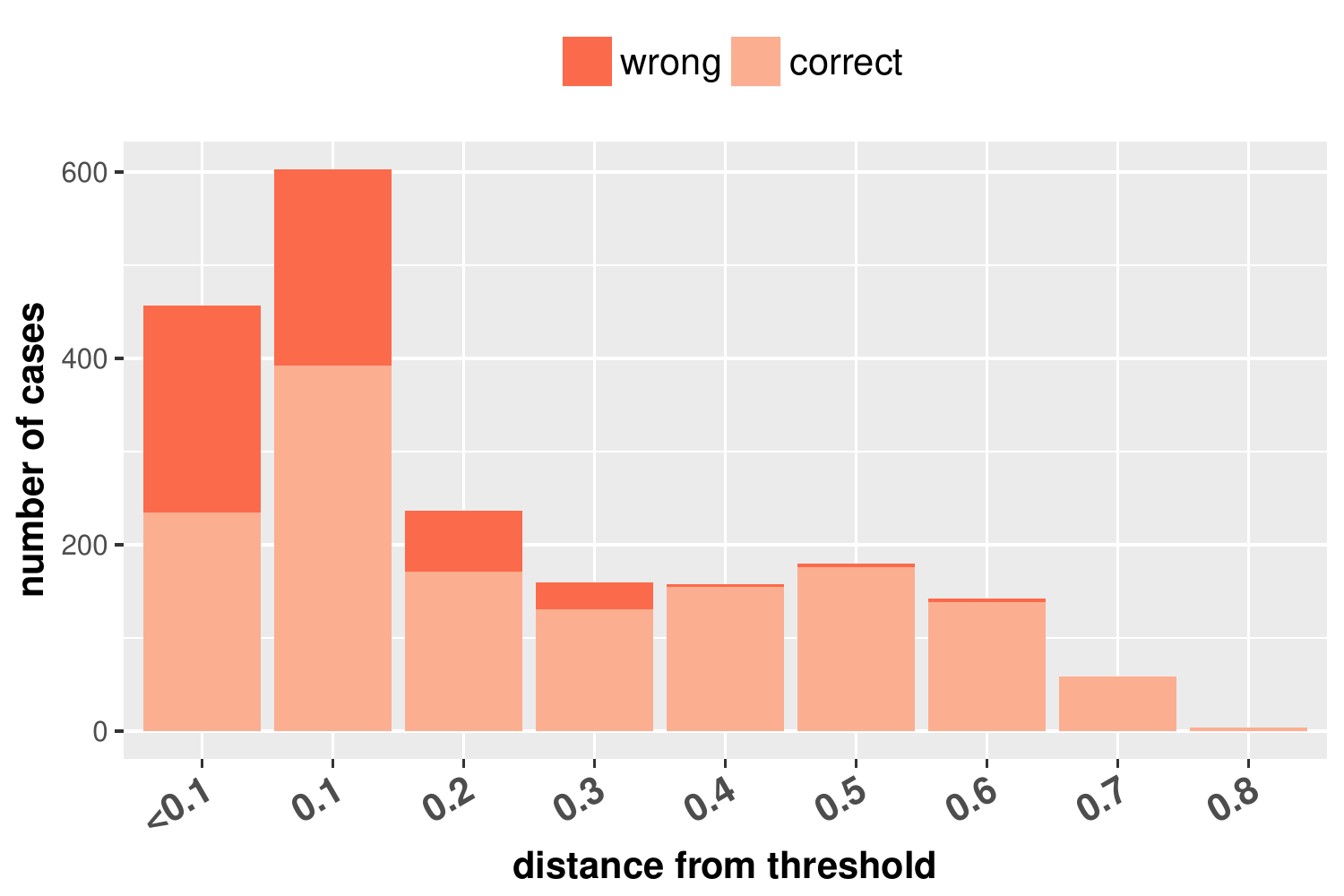}
\caption{\textbf{SET+POS-hard.} Correct and wrong predictions by best run of FiLM (73\% acc.) against distance from the threshold. Best viewed in color.}\label{fig:dhard_pred_glob} 
\end{figure}

\begin{figure*}[t!]
\centering
\begin{subfigure}[b]{0.45\linewidth}
\includegraphics[width=0.49\linewidth]{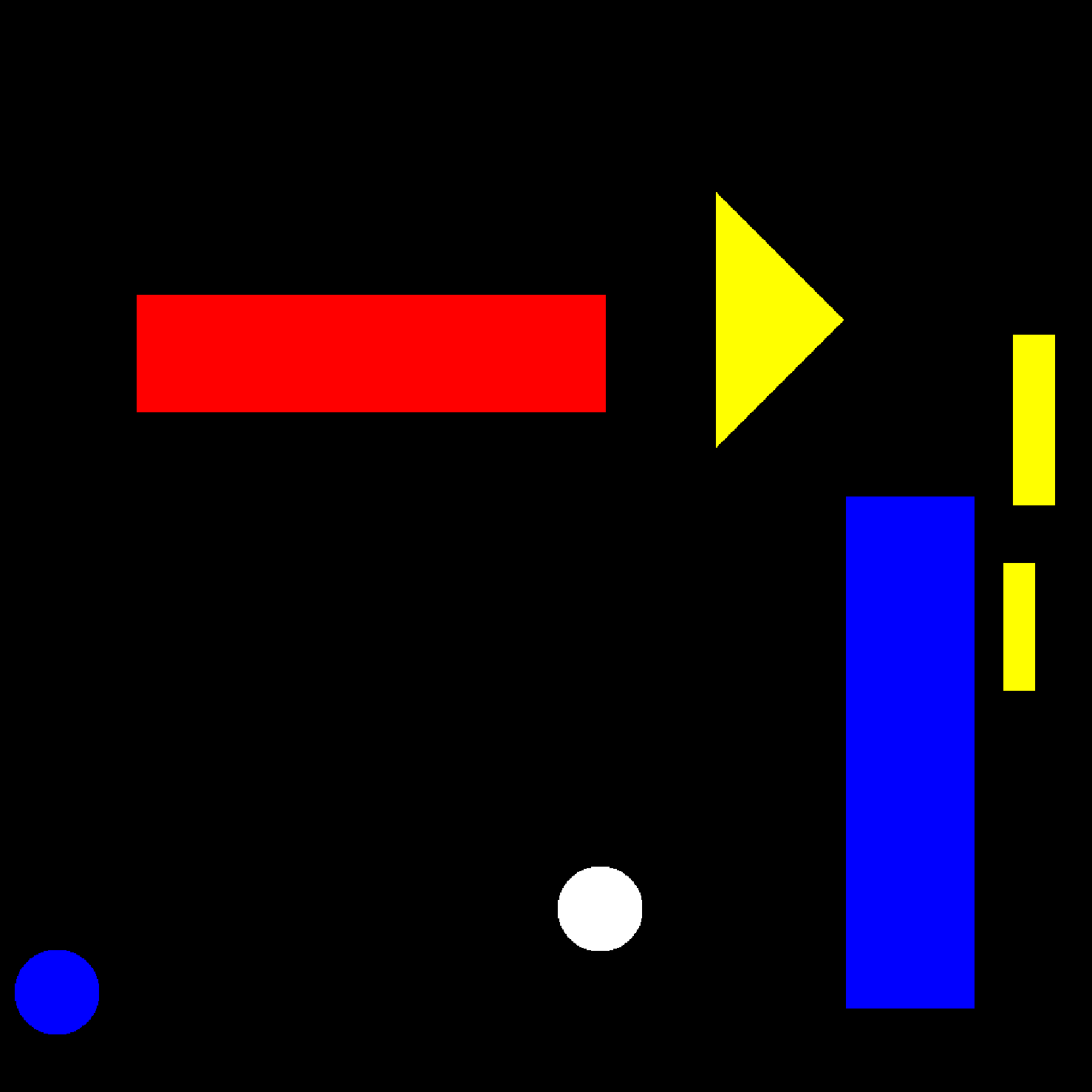}
\includegraphics[width=0.49\linewidth]{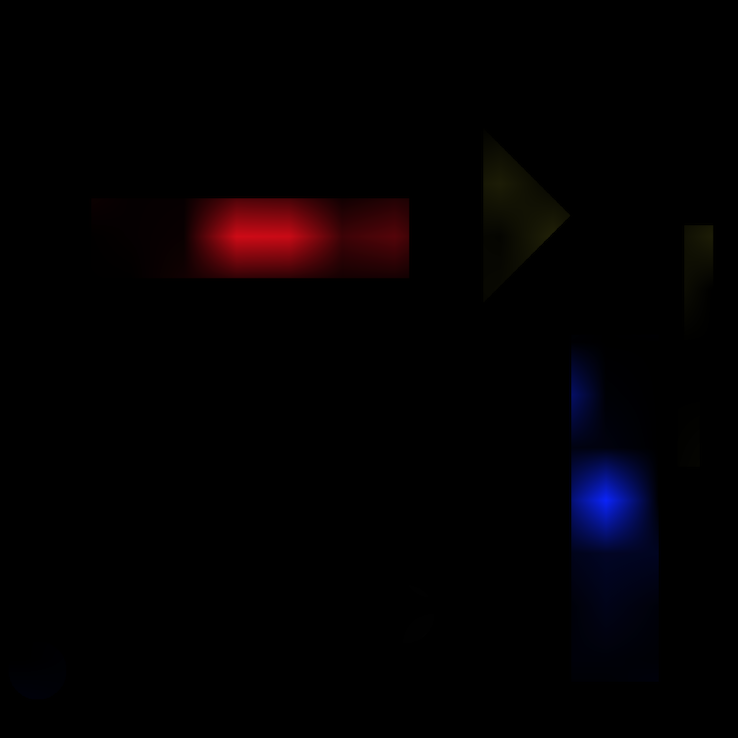}\hfill
\caption{\label{fig:fig1}}
\end{subfigure}%
\hfill
\begin{subfigure}[b]{0.45\linewidth}
\includegraphics[width=0.49\linewidth]{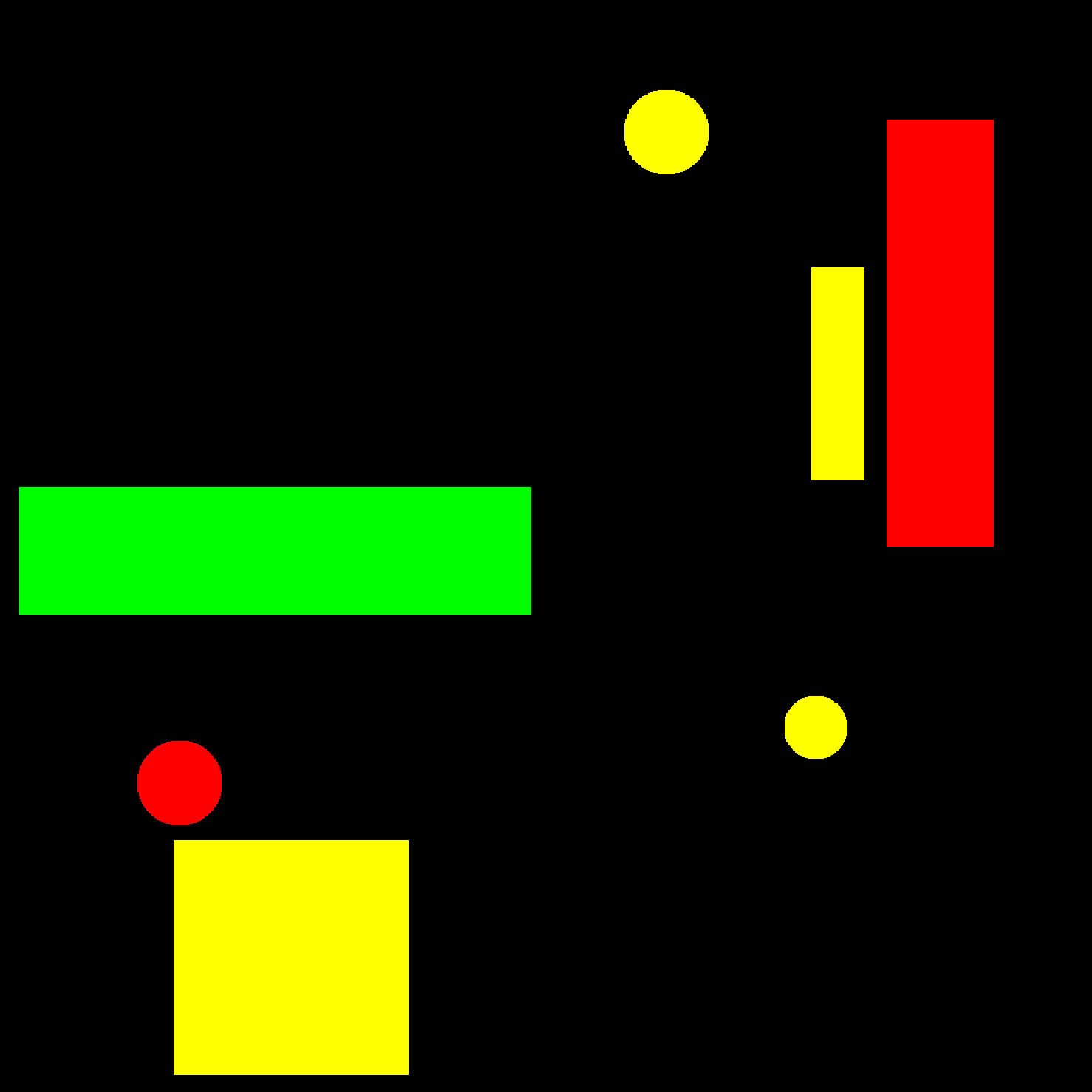}
\includegraphics[width=0.49\linewidth]{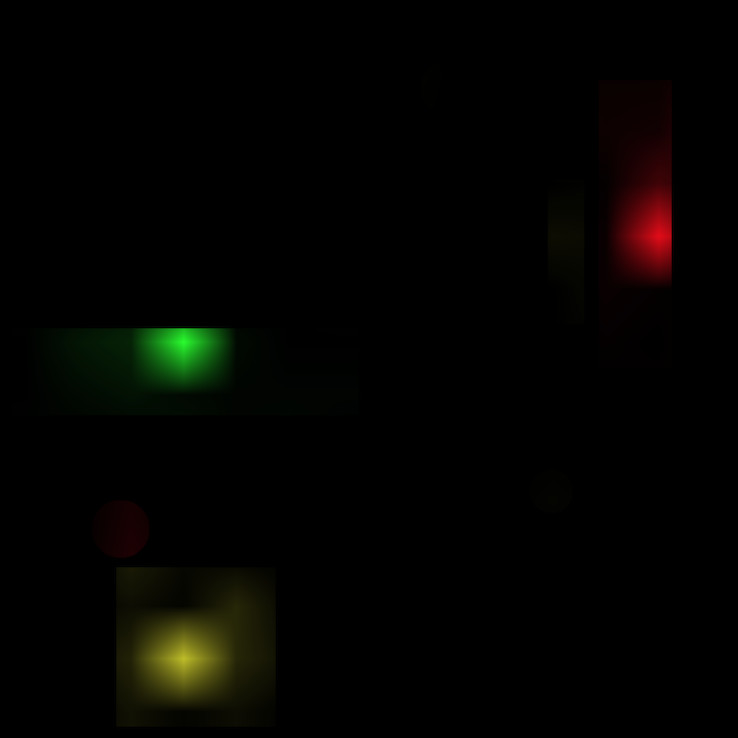}
\caption{\label{fig:fig2}}
\end{subfigure}%
\caption{\textbf{SET+POS-hard.} Two \emph{borderline} cases from the test set where FiLM's best run makes a correct (a) or wrong (b) prediction to `The red rectangle is a \emph{big} rectangle', which is \emph{true} in both scenes. The rightmost panels show the distribution of locations used by FiLM for its globally max-pooled features. Best viewed in color.}\label{fig:vis}
\end{figure*}

As for FiLM, its pattern of accuracy is overall more stable across sentence types, though big objects (big true, small false) are still significantly harder than small ones (small true, big false). Bigger objects are more challenging because by definition they are closer to the threshold (as can clearly be seen in Figure~\ref{fig:diagram}), which in turn means that they are more likely to be borderline cases. The plots in Figure~\ref{fig:distr_k_dhard} illustrate this effect, showing that objects that count as `big' are more likely to flip the truth value of a sentence due to the fuzziness of the threshold. FiLM's results are thus to be expected, and in fact encouraging: if a model is correctly learning to apply the threshold function to the reference set, most of the errors should involve borderline objects. This is confirmed: see Figure~\ref{fig:dhard_pred_glob}, where correct and wrong predictions by the best FiLM model are plotted against the (normalized) distance of the queried object from the threshold (83\% of the errors within the 2 leftmost columns correspond to objects that count as `big'). In Figure~\ref{fig:vis}, we report two \emph{borderline} cases randomly sampled from the test set where distance from the threshold is 0.09 (a) and 0.04 (b), respectively. For both scenes, the sentence `The red rectangle is a \emph{big} rectangle' is \emph{true}. However, FiLM's best run predicts the correct answer only in (a). By visualizing the distribution of locations responsible for FiLM's globally-pooled features fed to the classifier, we notice that, in both cases, features related to the queried object (the red rectangle) and objects in the reference set (rectangles) are highly activated (though with a bias toward larger objects), confirming the ability of FiLM to focus on language-relevant features. However, (high) activations are also assigned to \emph{unrelated} objects' features, particularly in (b). Though no conclusions can be drawn from these examples, they suggest that FiLM uses features related to several objects, in line with the \emph{relational} nature of the task.

Finally, the evidence that models obtain slightly different results across shapes (see Figure~\ref{fig:dhard_acc_shape}) could suggest that models learn different, shape-specific representations of `big' and `small'. The following analysis precisely aims at exploring this issue.

\subsection{Compositionality: Unseen Combinations}
\label{subsec:compo}


In our last analysis, we investigate whether the models learn an \emph{abstract} representation of GAs that can be compositionally applied to unseen adjective-noun combinations. We focus on the SET+POS task and extract a subset of the corresponding dataset in which each size adjective appears only with two nouns denoting two distinct shape types: `big (circle$\mid$rectangle)', `small (square$\mid$triangle)'. This subset of the data amounts to half of SET+POS and thus contains 10K datapoints, that we split into train (8K), val (1K), and test (1K). We refer to this test set as \emph{seen} (its adjective-noun combinations are seen in training). We then create a second, \emph{unseen} test set with 1K datapoints, where the adjectives appear with different nouns: `big (square$\mid$triangle)', `small (circle$\mid$rectangle)'.
We train all models three times using their best hyper-parameter configurations and evaluate them on both \emph{seen} and \emph{unseen}. If models learn an abstract representation of GAs that includes a variable for the noun they modify, we should observe no difference in performance between the two test sets.

As reported in Table~\ref{tab:comp}, this turns out not to be the case. While in the \emph{seen} test set all models obtain similar accuracies to those obtained in SET+POS, in \emph{unseen} their performance is not only well below chance level, but it is in fact the inverse of their results on \emph{seen}:  all instances which are predicted correctly in \emph{seen} are incorrectly predicted in \emph{unseen} (e.g, FiLM obtains $\sim$85\% on \emph{seen} and $\sim$15\% on \emph{unseen}). 
This suggests that the models learn a default strategy per noun rather than an abstract adjective representation that generalizes to unseen combinations. For example, the models predict \emph{true} any time the size of a circle or a rectangle exceeds a certain threshold, regardless of whether the noun is modified by `big' (\emph{seen} combinations) or `small' (\emph{unseen} combinations).

\begin{table}[h!]
\small
\resizebox{\columnwidth}{!}{
\begin{tabular}{|l|l|l|}
\hline
\multicolumn{1}{|c|}{\textbf{model}} & \multicolumn{1}{c|}{\textbf{avg \emph{seen} $\pm$ sd}} & \textbf{avg \emph{unseen} $\pm$ sd} \\ \hline
CNN+LSTM                             & 0.608 $\pm$ 0.01                                     & 0.4036 $\pm$ 0.003                \\
CNN+LSTM+SA                          & 0.7813 $\pm$ 0.009                                   & 0.235 $\pm$ 0.006                 \\
FiLM                                 & 0.8489 $\pm$ 0.014                                   & 0.153 $\pm$ 0.02                  \\ \hline
\textit{chance}                      & 0.5                                                  & 0.5                               \\ \hline
\end{tabular}
}
\caption{\textbf{Compositional task.} Results by the models.}\label{tab:comp}
\end{table}

%% file: sec_discussion.tex

\section{Discussion}
\label{se:disc}

We tackle the modeling of size GAs as a \emph{relational} problem in the domain of visual reasoning, and show~\cite[in contrast with][]{kuhnle2018clever} that FiLM is able to learn the function underlying the meaning of these expressions. However, none of the models develop an \emph{abstract} representation of GAs that can be applied \emph{compositionally}, an ability that even 4-year-olds master~\cite{barner2008}. This is in line with recent evidence showing that deep learning models do not rely on systematic compositional rules~\cite{baroni2019linguistic,bahdanau2019}.

An interesting open question, which we plan to explore in future work, is whether training models to jointly learn superlative, comparative, and positive GAs (similarly to how~\citet{N18-1039} did for quantities), or framing the task in a dialogue setting (as~\citet{monroe2017colors} did for colors) could lead to more compositional models. Moreover, it might be worth exploring whether equipping models with similar inductive biases as those leading speakers of any language to develop abstract, compositional representations of size adjectives is needed to properly handle these expressions. In parallel, the present work could be extended to other GAs and threshold functions.

In the long term, we aim to move to natural images. This requires world knowledge, a confounding factor intentionally abstracted away in synthetic data. Since children learn to exploit world knowledge after mastering the perceptual context~\cite{tribushinina2013adjective}, adopting an incremental approach might be promising.